%% file: main.tex
\documentclass[letterpaper]{article} 
\usepackage{aaai23}  
\usepackage{times}  
\usepackage{helvet}  
\usepackage{courier}  
\usepackage[hyphens]{url}  
\usepackage{graphicx} 
\usepackage{natbib}  
\usepackage{caption} 
\usepackage[ruled,linesnumbered]{algorithm2e}
\usepackage{booktabs}
\usepackage{multirow}
\usepackage{xcolor}

\usepackage{amsmath}

\usepackage{xspace}
\newcommand{\name}[0]{H-TSP\xspace}
\usepackage{amsfonts}

\title{H-TSP: Hierarchically Solving the Large-Scale Travelling Salesman Problem}
\author{
    Xuanhao Pan\textsuperscript{\rm 1},
    Yan Jin\textsuperscript{\rm 1}\thanks{Yan Jin is the corresponding author.},
    Yuandong Ding\textsuperscript{\rm 1},
    Mingxiao Feng\textsuperscript{\rm 2},
    Li Zhao\textsuperscript{\rm 3},
    Lei Song\textsuperscript{\rm 3},
    Jiang Bian\textsuperscript{\rm 3}
}
\affiliations {
    \textsuperscript{\rm 1} School of Computer Science, Huazhong University of Science and Technology, China,\\
    \textsuperscript{\rm 2} University of Science and Technology of China,  \\
    \textsuperscript{\rm 3} Microsoft Research Asia\\
    \{xhpan, jinyan, yuandong\}@hust.edu.cn, fmxustc@mail.ustc.edu.cn, \{lizo, lei.song, jiang.bian\}@microsoft.com
}

\begin{document}

\maketitle

\input{abstract}

\input{intro}

\input{relatedwork}

\input{problem}

\input{framework}

\input{training}

\input{experiment}

\section{Conclusion}
In this paper, we propose a hierarchical deep reinforcement learning framework for large-scale TSP, named H-TSP,
which solves TSP in a divide-and-conquer manner.
We test H-TSP on four datasets with different numbers of nodes, and the results show that H-TSP outperforms other SOTA baselines in terms of efficiency, 
while the solution quality of H-TSP remains comparable with the SOTA learning-based method.
The ablation study shows that the lower-level model significantly affects the performance of H-TSP,
and the solution quality can be further improved by replacing the lower-level model with SOTA LKH-3 solver during inference, allowing us to achieve a flexible trade-off between efficiency and solution quality.
Furthermore, we believe that the divide-and-conquer method can be generalized to other large-scale problems 
such as vehicle routing and job scheduling, and we leave these as our future works.

\appendix

\bibliography{aaai23}.

\bigskip

\end{document}

%% file: abstract.tex
\begin{abstract}
    
    We propose an end-to-end learning framework based on hierarchical reinforcement learning, called H-TSP, for addressing the large-scale Travelling Salesman Problem (TSP). The proposed H-TSP constructs a solution of a TSP instance starting from the scratch relying on two components: the upper-level policy chooses a small subset of nodes (up to 200 in our experiment) from all nodes that are to be traversed, while the lower-level policy takes the chosen nodes as input and outputs a tour connecting them to the existing partial route (initially only containing the depot). After jointly training the upper-level and lower-level policies, our approach can directly generate solutions for the given TSP instances without relying on any time-consuming search procedures. To demonstrate effectiveness of the proposed approach, we have conducted extensive experiments on randomly generated TSP instances with different numbers of nodes. We show that H-TSP can achieve comparable results (gap 3.42\% vs. 7.32\%) as SOTA search-based approaches, and more importantly, we reduce the time consumption up to two orders of magnitude (3.32s vs. 395.85s). To the best of our knowledge, H-TSP is the first end-to-end deep reinforcement learning approach that can scale to TSP instances of up to 10000 nodes. Although there are still gaps to SOTA results with respect to solution quality, we believe that H-TSP will be useful for practical applications, particularly those that are time-sensitive e.g., on-call routing and ride hailing service.
\end{abstract}

%% file: intro.tex
\section{Introduction}
Travelling salesman problem (TSP) is a well-known combinatorial optimization problem. It has been studied in operation research community for many years. The best exact solver Concorde~\citep{Applegate2009} requires 136 CPU years to find the optimal solution for an instance with 85,900 cities. Such computation time is unacceptable, so many heuristics have been proposed to obtain near-optimal solutions for problems arising in practice. For instance, one of the leading heuristics algorithms LKH3~\citep{Helsgaun2017} can handle TSP instances with millions of cities. However, these algorithms consist of hand-crafted rules that are specific to TSP. More importantly, heuristics rely on iterative search and are also time-consuming for large-scale instances. This limits their applications in scenarios that are time sensitive, e.g. on-call routing~\cite{GHIANI20031} and ride hailing service~\cite{Xu2018}.

To deal with such time sensitive applications, learning-based approaches for TSP are expected to be used as they are very efficient during inference. They can learn useful patterns from massive data during training which can generalize to unseen instances. Moreover, they do not rely on problem specific knowledge, hence can be extended to handle similar problems. Because of these advantages, there have been a soaring number of studies to solve TSP in recent years, for instance~\citep{NIPS2015_29921001,Nowak2017,Kool2019,Kwon2020,Fu2020}.

Depending on how solutions are constructed, learning based approaches can be classified into two categories: iterative and constructive. For iterative approaches, a feasible solution shall be constructed at the beginning, either randomly or using some heuristic methods. The main step is to learn a model that knows how to improve a feasible solution. Finally, the model will be applied to keep improving the current solution until a termination condition is met. On the other hand, constructive approaches start from the initial city and learn which city to go at each step. While these approaches have achieved competitive performance in relatively small TSP instances (less than 1000 cities), they cannot be extended to deal with large-scale TSP easily. One exception is~\citep{Fu2020} which can achieve solutions close enough to LKH3 solutions (gap $<5\%$) but in relatively shorter time. However, in~\citep{Fu2020} a Monte Carlo tree search procedure is required to improve solutions constantly, which is still time-consuming. According to the experiment results in~\citep{Fu2020}, for one TSP instance with 10,000 cities, it costs around 11 minutes until it finds a solution comparable with LKH3.

We noticed that the iterative approaches are time-consuming on training and inference of large-scale TSP, since they require to constantly improve feasible solutions to obtain solutions of high quality by learning a proper operator from some domain-specific heuristic operators. And the constructive approaches are efficient on inference, but due to their action spaces grow linearly in the number of cities, their training procedures would easily run out of memory or time before converging to some near-optimal solutions.

To overcome these challenges, we propose a constructive approach based on hierarchical reinforcement learning (\name for short), that can obtain comparable results as SOTA approaches but using much less time up to two orders of magnitude. 
Starting from a partial route (initially only containing the depot), our \name approach decomposes solutions construction into two steps: Firstly, we shall choose a relatively small subset of cities from all remaining cities that are to be visited; Secondly, we solve a small open loop TSP instance that only contains the chosen cities. Once a tour for the chosen cities is obtained, it will be merged into the existing partial route. Such a procedure will continue until all cities have been visited and a feasible solution is obtained. Correspondingly, we devise two policies for the two steps: one is to choose candidate cities to be traversed, while the other one decides in which order these cities will be visited. These two policies are trained jointly using reinforcement learning algorithms so that we obtain an end-to-end algorithm for solving TSP. Our \name solves TSP in a divide-and-conquer manner so it can scale to large-scale TSP easily. 

The contributions are summarized as follows:

\begin{itemize}
    \item We propose an effective hierarchical framework, named \name , mainly for addressing the large-scale TSP. Using a divide-and-conquer strategy, \name is the first end-to-end approach that can scale to TSP with up to 10,000 nodes, to the best of our knowledge; 
    \item We have conducted extensive experiments to show effectiveness of \name. As demonstrated by experiment results, \name can achieve comparable results of instances with up to 10,000 nodes as SOTA search-based approaches, while the time spent are less than 4 seconds. Notice that the computation time is reduced considerably, up to two orders of magnitude. Therefore, our approach will be particularly useful for time sensitive applications;
    \item Decomposition is a common technique to solve large-scale combinatorial optimization problems~\cite{mariescu-istodorSolvingLargescaleTSP2021}. We believe that the framework proposed in this paper has potentials to be extended to other problems. We leave it as our future work to solve other large-scale optimization problems following the divide-and-conquer framework.
\end{itemize}

%% file: relatedwork.tex
\section{Related Work}
Due to the importance of TSP in research and industry, there have been lots of studies on this topic. Instead of giving a comprehensive overview, we mainly focus on related studies. We shall refer interested readers to~\citet{Taillard2019} and \citet{Guo2019,Bengio2021} for overviews of heuristic approaches and other machine learning approaches, respectively.

\subsection{Traditional TSP algorithms}
Traditional TSP algorithms can be roughly classified into two categories, exact algorithms and heuristic algorithms.
Concorde~\cite{applegate2007traveling} is one of the best exact solvers. It models TSP as mixed-integer programming problems, then uses a branch-and-cut algorithm~\cite{padberg1991branch} to solve it.
LKH-3~\cite{Helsgaun2017} is a SOTA heuristic for solving TSP. It adopts the idea of local search and uses the $k$-opt operators and an $\alpha$-nearness measure to reduce the search space. However, to obtain high quality solutions, LKH-3 often takes hours or longer to terminate when solving TSP with tens of thousands nodes. 

\subsection{Learning-based TSP algorithms}
Depending on how solutions are constructed, learning-based algorithms can be categorized into two main categories: constructive-based methods and search-based methods.

\paragraph{Constructive-based methods} 
Pointer network~\cite{vinyals2015pointer} is known as the first end-to-end method that incrementally generates solutions of TSP from the scratch. The backbone model is a Recurrent Neural Network, which is trained in a supervised manner. Differently, Bello~\cite{bello2016neural} improved the performance of pointer network by training it using reinforcement learning, hence can achieve policies with better generalization. 
Inspired by the success of Transformer~\citep{Vaswani2017} in many fields, such an architecture has also been extended to deal with TSP~\citep{Kool2019,Kwon2020}. Additionally, \citep{Kool2019} introduces a simple baseline based on a deterministic greedy roll-out to train the model using REINFORCE~\citep{Willia1992}. The work in~\citep{Kwon2020} further exploits the symmetries of TSP solutions, from which diverse roll-outs can be derived so that a more efficient baseline than~\citep{Kool2019} can be obtained. However, most of these works focus on solving TSP with cities no more than 100, except that~\citep{Dai2017} considers instances with up to 1,200 cities.

\paragraph{Search-based methods} 
In~\cite{pmlr-v129-costa20a}, a method called DRL-2opt is proposed that uses DRL algorithms to train a policy. Such a policy will select suitable 2-opt operators to continuously improve the current solution. Another approach called VSR-LKH in Zheng~\cite{zhengCombiningReinforcementLearning2021a} is proposed. VSR-LKH can be seen as a variant of LKH solver. Differently, it replaces $\alpha$-nearness in LKH by a Q-value table that is obtained by 
running reinforcement learning algorithms.
Similar as our approach, an approach called Att-GCN+MCTS is introduced in~\cite{Fu2020} that utilizes decomposition mechanism to solve large-scale TSP. It consists of three stages: Firstly, a supervised model is trained that can generate heat maps for small TSP instances;
Secondly, techniques like graph sampling, graph converting, and heat map merging are introduced that can generate heat maps of large-scale TSP from smaller ones;
Finally, a Monte Carlo tree search procedure is introduced to search for a good solution guided by the heat map.
Search-based methods can often obtain high-quality solutions if given enough time. This limits their applications in scenarios that are time sensitive.

It is worth to mention another related work~\cite{NEURIPS2021_c6a01432} that relies on Hierarchical Reinforcement Learning (HRL) to solve large-scale Dynamic Pickup and Delivery Problems (DPDPs) in practice. DPDPs can be seen as a variant of TSP, where nodes are unknown a priori but will be released periodically. The objective is to assign these nodes to proper routes in nearly real time which minimize the total traverse cost. To address the problem, a hierarchical framework is introduced, where the upper-level policy dynamically partitions the problem into sub-problems, while the lower-level policy tries to solve each sub-problem efficiently. The main difference between this method and \name is that their method aims to partition a dynamic problem into multiple static sub-problems at the temporal level, while \name aims to decompose a large-scale TSP problem into a set of small sub-problems at the spatial level.

%% file: problem.tex
\section{Problem Definition}
In this paper, we focus on two-dimensional Euclidean TSP.
Let $G=(V,E)$ denote an undirected graph,
where $V=\left\{v_i, | 1 \leq i \leq N \right\}$ represents the set of nodes, 
$E=\left\{e_{i,j} | 1 \leq i,j \leq N \right\}$ is the set of edges, and
$N$ denotes the number of nodes.
For every edge $e_{ij}$, define $cost(i,j)$ as its traverse cost, namely, the distance between $i$ and $j$.
We define a special node $v_d \in V$ representing the depot node where the salesman starts from and ends at.
A feasible solution of a TSP instance is defined as a Hamiltonian cycle that visits all the nodes in $V$ exactly once.
Our goal is to minimize the total cost of the solution route $\tau$, which can be written as $L(\tau)$, shown in Eq.(\ref{eq_Ltau}).
\begin{equation}
   \label{eq_Ltau}
    L(\tau) = \sum_{i=1}^{N-1} cost(\tau_{i},\tau_{i+1}) + cost(\tau_{N},\tau_{1})
\end{equation}
Where $\tau_{i}$ is the $i$-th node in the route.
Without loss of generality, we assume all coordinates are in $[0,1]$.

\subsection{Sub-problem definition}\label{subsec:sub-problem-definition}
As mentioned before, our hierarchical method decomposes TSP into small sub-problems at the spatial level.
In order to facilitate sub-solutions merging, we define our sub-problem as a variant of TSP,
named open-loop TSP with fixed endpoints~\cite{Papadimitriou1977}.
Given an undirected graph $G=(V,E)$, the open-loop TSP has two special nodes $v_{s}$ and $v_{t}$ in $V$ representing the source node and target node, respectively.
A feasible solution of a open-loop TSP is no longer a cycle but a path, which starts from $v_{s}$,
visits all other nodes exactly once and ends in $v_{t}$.
It is easy to see that solutions of two open-loop TSPs can be merged to a close-loop path. 
The two endpoints in the sub-problem need to be fixed and specified in advance,
if not, the sub-solutions will have arbitrary endpoints, 
which further leads to a poor combined solution.

%% file: framework.tex
\section{The Hierarchical Framework}
This paper proposes a Deep Reinforcement Learning (DRL) based hierarchical framework (denoted as H-TSP) to solve the large-scale TSP.
Following the divide-and-conquer approach, H-TSP contains policies/models in two levels, which are responsible for generating sub-problems and solving sub-problems, respectively.

The entire procedure of H-TSP is summarized in Algorithm~\ref{alg:hierarchical-tsp}. It starts with an initial solution containing the depot, and then inserts the node nearest to it as two fixed endpoints of the first sub-problem. The upper-level policy is responsible for decomposing the original problem and merging sub-solutions from the lower-level policy.
As decomposition will inevitably downgrade quality of final solutions, to alleviate this we let the upper-level policy learn to generate a decomposition strategy in an adaptive and dynamic manner. On the other hand, once a sub-problem is identified, it will be handed over to the lower-level policy to solve it as a open-loop TSP. Its solution will then be passed to the upper-level policy to merge into the existing partial route. 
\vspace{-10pt}
\begin{algorithm}
    \caption{Hierarchical TSP Algorithm}\label{alg:hierarchical-tsp}
    \KwIn{TSP instance $V=\{v_1, v_2, \cdots, v_N \}$, initial solution $\tau_{init} = \{v_{d} \}$}
    \KwOut{Solution route $\tau=\{ \tau_1, \tau_2, \cdots, \tau_N \}$}
    $\tau \gets \tau_{init}$, the nearest node $v$ of $v_d$ \;
    \While{$\text{len}(\tau) < N$}{
        $SubProb \gets GenerateSubProb(V, \tau)$ \;
        $SubSol \gets SolveSubProb(SubProb)$ \;
        $\tau \gets MergeSubSol(SubSol, \tau)$ \;
    }
    return $\tau$
\end{algorithm}
\vspace{-10pt}

\subsection{Upper-level Model}
As mentioned before, the upper-level model is to decompose the large-scale TSP so that they can be solved efficiently without significantly downgrading quality of final solutions. To achieve this, we conduct decomposition in an adaptive and dynamic manner. This differentiates from the existing decomposition-based approach in~\cite{Fu2020}, where all sub-problems are pre-generated before merging them into the final solution. By interleaving decomposition and merging, the upper-level model can learn an adaptive policy that can make its best decision based on current partial solution and distribution of remaining nodes.

\paragraph{A Scalable Encoder}
One of the key obstacles in solving the large-scale TSP with DRL is to encode a large number of edges in the graph.
For achieving a scalable encoder, inspired by a technique used in 3D point cloud projection~\cite{8954311}, we propose a Pixel Encoder to encode the graph as pixels. The idea is to convert point clouds into a pseudo-image, which are nodes of a TSP instance in our case.

As a first step the 2D space is discretized into an evenly spaced $H\times W$ grid, creating a set of pixels.
Then the nodes are divided into different clusters based on the grid they are on.
We augment features of each node with a vector $(x_a, y_a, \Delta x_g, \Delta y_g, \Delta x_c, \Delta y_c, x_{pre}, y_{pre}, x_{nxt}, y_{nxt}, m_{select})$,
where $(x_a, y_a)$ is the absolute coordinate of the node, and
$(\Delta x_g, \Delta y_g)$ and $(\Delta x_c, \Delta y_c)$ are the relative coordinates to the gird center and the node cluster center, respectively.
If the node has been visited, we let $(x_{pre}, y_{pre}, x_{nxt}, y_{nxt})$ denote coordinates of its neighbors on the partial route, otherwise they are 0.
The boolean variable $m_{select}$ indicates whether this node has been visited or not.

For a TSP instance with $N$ nodes, we have a tensor of size $(N, D)$, where $D=11$ is the number of features.
Then this tensor is processed by a linear layer to generate a $(N, C)$ sized high dimensional tensor.
According to the divided clusters, we use a max operation over the $C$ dimension to get the feature of each grid,
and we use zero padding for the empty grids. 
The combination of each grid forms a pseudo-image: a tensor of size $(H, W, C)$.
This pseudo-image can be further processed by a convolutional neural network (CNN),
resulting in a embedding vector of the whole TSP instance for the DRL model.

The DRL model follows the actor-critic architecture,
there is a policy head for the policy function and a value head for the state value function.
Both of the two heads are composed of fully connected layers and activation functions.

\paragraph{Sub-problem generation and merging}
The action space of our upper-level policy is continuous with 2 dimensions, each of which denotes coordination of a point in the grid. We illustrate in this paragraph how a sub-problem is generated and merged given an upper-level policy. The procedure is also depicted in Algorithm~\ref{alg:sub-problem-generation}.
For a given action $Coord_{pred}$, let $v_c$ be the node closest to $Coord_{pred}$ that has not been visited yet.
We further let $v_b$ be the node closed to $v_c$ that has already been visited.
Then, we keep expanding the sub-problem by selecting nodes that have not been visited from neighbors of $v_{b}$ based on a $k$-Nearest Neighbor ($k$-NN) graph, until size of the sub-problem reaches $maxNum$ or all nodes are visited or selected. Intuitively, the $k$-NN graph is implemented by associating each node a set of $k$ closest nodes and the sub-problem expansion will follow a breadth-first search on the $k$-NN graph.
Finally, we enrich the sub-problem ($SelectFragment$) with a fragment of visited nodes centering at $v_{b}$ so that the resulting sub-problem has nodes not greater than $subLength$. In this way, we break the existing partial route to obtain a path with two endpoints, while after solving the sub-problem as a open-loop TSP, we would obtain another path with two endpoints. 

\begin{algorithm}[t]
    \caption{Sub-problem Generation}\label{alg:sub-problem-generation}
    \KwIn{
        $k$-NN graph $G_{kNN}=(V,E)$, the partial solution at step $t$ $\tau_t = \left\{v_t^1, v_t^2, \cdots\right\}$,
        length of the sub-problem $subLength$, maximum number of unvisited nodes $maxNum$,
        upper layer model $UpperModel$}
    \KwOut{Sub-problem $P=\left\{v_s^1, v_s^2, \cdots, v_s^{subLength}\right\}$, two endpoints $v_{s}, v_{t} \in P$}
    $P \gets \emptyset, S_v \gets \tau_t, Q_{new} \gets Deque()$\;
    $UpperModel$ inputs $G_{kNN}$ and $\tau_t$, outputs $Coord_{pred}$ \;
    $v_{c}$ is the unvisited node closest to $Coord_{pred}$\;
    $v_{b}$ is the visited node closest to $v_{c}$ \;
    Push $v_{b}$ to the end of $Q_{new}$ \;
    \While{$\text{len}(P) \leq maxNum$ and $Q_{new}$ \textup{is not empty}}{
        $v_i \gets PopFront(Q_{new})$ \;
        \For{$v_j \in N_{G_{kNN}}(v_i)$ and $v_j \notin S_v$}{
            Push $v_{j}$ to the end of $Q_{new}$ \;
            Add $v_j$ into $S_v$ and $P$ \;
        }
    }
    $oldLength = subLength-\text{len}(P)$ \;
    $P_t \gets SelectFragment(\tau_t, v_b, oldLength)$ \;
    $P \gets P\cup P_t$ \;
    $v_{s}, v_{t} \gets SetEndpoints(P_t)$ \;
    return $P, v_{s}, v_{t}$\;
\end{algorithm}


\paragraph{Markov Decision Process}
Now we shall be able to introduce underlying MDPs of upper-level policies formally in this paragraph.
Let $M_G=\langle S, A, P, R, \gamma\rangle$ denote an MDP modelling a given TSP instance $G=(V, E)$, where 
\begin{itemize}
    \item $S$ is the set of all states containing all possible path fragments $\tau$ of $G$;
    \item $A=[0,1]\times [0,1]$ is the set of all actions containing all points in a unit grid;
    \item $P:S\times A\rightarrow S$ is a deterministic transition function given both upper-level and lower-lever policies;
    \item $R:S\times A\times S\rightarrow \mathbb{R}$ is the reward function defined by $R(\tau, a, \tau')=L(\tau)-L(\tau')$;
    \item $\gamma$ is the discount factor, which we set to 1 in our experiments. 
\end{itemize}

\subsection{Lower-level Model}
The lower-level model is trained for solving open-loop TSPs with fixed endpoints generated by the upper-level model. As lower-level policies will be launched for many times during training and interference, its performance will have a significant impact on the performance of our approach. Fortunately, there have been many end-to-end approaches that can solve relatively small-scale TSPs effectively and efficiently~\cite{Kool2019,Kwon2020}. We adopt main ideas of these approaches to devise an efficient lower-level policy, which we will briefly illustrate in this section. 

\paragraph{Neural Network}
The underlying neural network of our lower-level model is a Transformer network, which has been widely used in the fields of natural language processing and computer vision in recent years. It consists of a Multi-Head Attention and a Multi-Layer Perceptron layer,
with a mask mechanism to remove all invalid actions.

Our neural network follows the encoder-decoder structure, where the encoder uses self-attention layers to encode the input node sequence,
while the decoder outputs a sequence of nodes in an auto-regressive manner.
In approaches presented in~\cite{Kool2019,Kwon2020}, the following context is used as input of the encoder,
\begin{equation}
    q_{context} = q_{graph} + q_{first} + q_{last}
    \label{eq:tsp-context}
\end{equation}
where $q_{graph}$, $q_{first}$, and $q_{last}$ represent the feature vectors of the whole graph,
the first node and the last node of the current partial solution, respectively.
While enough for TSPs, it is inadequate for open-loop TSPs where we shall keep in mind that there are two fixed endpoints.
Therefore, we add two more vectors to encode the features of the two endpoints, namely, input of our encoder is a context vector defined as follows:
\begin{equation}
    q_{context} = q_{graph} + q_{first} + q_{last} + q_{source} + q_{target}
    \label{eq:sub-problem-context}
\end{equation}

The POMO approach introduced in~\cite{Kwon2020} takes advantage of the symmetry property of TSPs, which improves its performance considerably. Although open-loop TSPs do not have the same symmetry property due to existing of fixed endpoints, we can achieve such a symmetry property easily as follows: During the node selection, all nodes except the endpoints will be treated as in TSP without any constraint. Whenever an endpoint is chosen, we let the other one be chosen automatically. The final solution of the origin open-loop TSP is obtained by removing the redundant edge between the two endpoints.

\paragraph{Markov Decision Process}
The underlying MDPs of the lower-level policies can be defined similarly as in~\cite{Kool2019,Kwon2020}, where
\begin{itemize}
\item \textbf{States:} The states contain all possible contexts defined as in Eq.~(\ref{eq:sub-problem-context});
\item \textbf{Action:} The actions contain all nodes in a TSP, with dynamic masks to remove nodes that have been visited;
\item \textbf{Rewards:} A reward equating the negative cost of a route is assigned whenever a state corresponding to a feasible solution is encountered; otherwise the reward is 0.
\end{itemize}

%% file: training.tex
\section{Training}

The proposed framework are trained by a hierarchical DRL algorithm.
More specifically, the two levels of models are trained with DRL jointly.

\subsection{Upper-level Model}
The upper-level model is trained by the known Proximal Policy Optimization (PPO)~\cite{schulman2017proximal} algorithm, 
which is one of SOTA DRL algorithms based on an actor-critic architecture. It learns a stochastic policy by minimizing the following clipped objective function:

\begin{equation}
   \vspace{-3pt}
    L(\theta)=\hat{\mathbb{E}}_{t}\left[\min \left(r_{t}(\theta) \hat{A}_{t}, \operatorname{clip}\left(r_{t}(\theta), 1-\epsilon, 1+\epsilon\right) \hat{A}_{t}\right)\right]
\end{equation}
where $r_{t}(\theta)=\frac{\pi_{\theta}\left(a_{t} \mid s_{t}\right)}{\pi_{\theta_{\text {old }}}\left(a_{t} \mid s_{t}\right)}$ denotes the probability ratio of two policies,
$\hat{A}_t$ denotes the advantage function, $\epsilon$ is a hyperparameter controlling the clipping range.
The advantage function represents the advantage of the current policy over the old policy,
here we use the Generalized Advantage Estimator (GAE) to compute the advantage.

\begin{equation}
    \hat{A}_{t}=\sum_{l=1}^{\infty}(\gamma \lambda)^{l}\left(r_{t}+\gamma \hat{V}\left(s_{t+l+1}\right)-\hat{V}\left(s_{t+l}\right)\right)
\vspace{-5pt}
\end{equation}
where $r_t$ is the reward at time $t$, $\hat{V}$ is the state value function,
$\gamma$ is the discount factor, and $\lambda$ is the hyper-parameter that controls the compromise between bias and variance of the estimated advantage.

Besides the policy loss, we also add the value loss and entropy loss:
\begin{eqnarray}
    L^{\hat{V}}(\theta) = \hat{\mathbb{E}}_{t}\left[\hat{V}-\hat{V}_{\theta}\right]^2 \\
    L^{E}(\theta) = \hat{\mathbb{E}}_{t} \left[\pi_{\theta}\left(a \mid s\right) \log \pi_{\theta}\left(a \mid s\right)\right]
\end{eqnarray}

The total loss of upper-level model is:
\begin{equation}
    L^{UPPER}(\theta) = L(\theta) + L^{\hat{V}}(\theta) - \lambda_{e}L^{E}(\theta)
\end{equation}
where $\lambda_{e}$ is the weight of entropy loss for balancing the policy's exploration and exploitation.

\subsection{Lower-level Model}
The lower-level model is an end-to-end model for solving open-loop TSPs with relatively a small amount of nodes.
It is trained by the classic REINFORCE~\cite{williamsSimpleStatisticalGradientfollowing1992} algorithm with a shared baseline, as in~\cite{Kool2019,Kwon2020}.
The REINFORCE algorithm collects experience by Monte Carlo sampling and the policy gradient is computed as follows:

\begin{equation}
\vspace{-5pt}
    \begin{aligned}
        \nabla_{\theta} J(\theta) & =\mathbb{E}_{\pi_{\theta}}\left[\nabla_{\theta} \log \pi_{\theta}(\tau \mid s) A^{\pi_{\theta}}(\tau)\right]                             \\
                                  & \approx \frac{1}{N} \sum_{i=1}^{N}\left(R\left(\tau^{i}\right)-b(s)\right) \nabla_{\theta} \log \pi_{\theta}\left(\tau^{i} \mid s\right)
    \end{aligned}
\end{equation}
where $\tau$ denotes a trajectory, namely, a feasible solution of a TSP instance.
The reward $R(\tau^{i})=-L(\tau^i)$ is defined as the negative cost of $\tau^i$.
The shared baseline $b(s)$ is used to reduce the variance and improve the training stability, which is obtained by averaging the return of a set of trajectories that are generated from the same instance:
\begin{equation}
    b(s) = \frac{1}{N} \sum_{i=1}^{N}\left(R\left(\tau^{i}\right)\right)
\end{equation}

\subsection{Joint Training}
In order to improve the performance of the upper-level and lower-level models, we adopt a joint training strategy.
Specifically, the current lower-level policy will be used to collect trajectories for training the upper-level model,
and in the meanwhile, sub-problems generated by the upper-level policy will in turn be stored to train the lower-level model.
By such an interleaving training procedure, policies in two levels can receive instant feedback from each other, hence make the learning of a cooperative policy possible. 

As mentioned before, solution quality of lower-level policies has a significant impact on the final solution. If we start from a random lower-level policy, the upper-level policy would receive much misleading feedback making its training hard to converge. To alleviate it, we introduce a warm-up stage for the lower-level model by pre-training it with sub-problems randomly generated from the original TSP. According to our experiment, such a warm-up stage will accelerate convergence and make the training more stable.

%% file: experiment.tex
\section{Experiments}

This paper focuses on the large-scale TSP problem. To demonstrate how our approach works, we adopt four datasets to evaluate it. The four datasets contains TSP instances with problem sizes of 1000, 2000, 5000, and 10000 nodes, denoted as Random1000, Random2000, Random5000, and Random10000, respectively.
To make experiment results comparable, Random1000 and Random10000 contain the same instances used by Fu et al. in their work~\cite{Fu2020},
while instances in Random2000 and Random5000 are generated with nodes that are uniformly distributed in a unit square, in line with existing approaches.
Each dataset contains 16 TSP instances except Random1000, which contains 128 instances.
All our experiment results were obtained on a machine with an NVIDIA® Tesla V100 (16GB) GPU and Intel(R) Xeon(R) Platinum CPU.

\textbf{Hyperparameters setting} 
 The upper-level model consists of a pixel encoder and a DRL agent model.
We use a 3-layer CNN for the pixel encoder with 16, 32, 32 channels respectively, and it outputs a 128 dimensional feature vector.
And our DRL model with actor-critic architecture consists of a actor network and a critic network,
each of them is a 4-layer MLP.
The lower-level model follows the encoder-decoder structure,
there is a 12-layer self-attention encoder and a 2-layer context-attention decoder.
Most of the embedding dimension in the neural network is set to 128
except for the CNN layers, the first encoding layer and the outputting layer.
During training, we use the AdamW optimizer with a learning rate of 1e-4 and a weight decay of 1e-6.
For the sub-problem generation stage, we set $k=40$ for the $k$-nearest neighbor
and set the sub-problem length as 200 and the maximum number of new nodes in sub-problem as 190.
The lower-level model is trained for 500 epochs in the warm-up stage,
and the joint training stage takes 500, 1000, 1500, 2000 epochs respectively for different datasets.


\textbf{Baselines} We apply the following six SODA TSP solvers for comparison. \textbf{(1) Concorde}~\cite{applegate2007traveling} is one of the SOTA exact solvers for TSP. \textbf{(2) LKH-3}~\cite{Helsgaun2017} is one of the SOTA heuristic solvers for TSP.
\textbf{(3) OR-Tools}\footnote{OR-Tools. https://developers.google.com/optimization/} is an operational problem solver released by Google.
It has a wide range of applications and can solve a variety of combinatorial optimization problems such as TSP, VRP, and packing problems.
\textbf{(4) POMO}~\cite{Kwon2020} is an end-to-end DRL-based TSP algorithm,
and its performance is comparable to the SOTA methods.
\textbf{(5) DRL-2opt}~\cite{pmlr-v129-costa20a} is a search-based DRL algorithm, and one of the SOTA method in this category.
\textbf{(6) Att-GCN+MCTS}~\cite{Fu2020} is a novel method that combines GCN model trained with supervised learning
and MCTS searching to solve large-scale TSP. It can solve TSP instances with up to 10,000 nodes at the cost of a long searching time.

\subsection{Comparative Study}

We conduct comparative study on four randomly distributed TSP datasets.
The experimental results are shown in Table~\ref{tab:exp-1000-2000}.
The time in tables is the average time required for each instance, 
and Concorde is not tested on Random5000 and Random10000 as the exact solver takes too much time.
\name achieves comparable results to the SOTA methods in terms of solution quality.
The length of route generated by H-TSP is close to the search-based approach Att-GCN+MCTS
and much shorter than the two DRL-based algorithms.
Moreover, H-TSP outperforms all baselines in terms of efficiency, 
the computation time of H-TSP is one to two magnitude less than the baseline algorithms.
The high efficiency indicates that H-TSP has significant potential in many real-world scenarios
that require solving large-scale TSP in a short time even real-time.
It is worth noting that POMO and DRL-2opt perform poorly in all experiments,
because they cannot be trained directly on large-scale graphs,
and models trained on small-scale graphs cannot be generalized to large-scale graphs as well.
It is precisely the shortcomings of these two types of methods that motivate us to propose a new approach.
For the three baselines, LKH-3, OR-Tools and Att-GCN+MCTS, that have better solution quality than \name,
we have attempted to limit their searching time to the same magnitude as \name.
However, these methods require at least ten times as much time as \name in order to generate a feasible solution.

\begin{table}\small
    \centering
    \caption{Comparisons with seven solvers on large scale TSP}
    \vspace{-5pt}
    \label{tab:exp-1000-2000}
    \setlength{\tabcolsep}{1.5mm}{
    \begin{tabular}{@{}lllllll@{}}
        \toprule
        \multirow{3}{*}{Algorithm} & \multicolumn{3}{l}{Random1000} & \multicolumn{3}{l}{Random2000}                                              \\ \cmidrule(l){2-7}
                                   & Length                         & Gap                            & Time      & Length & Gap      & Time       \\ 
                                   &                           & (\%)                            & (s)      &   & (\%)       & (s)       \\ \midrule
        Concorde                   & 23.12                          & 0.00                        & 487.89 & 32.48  & 0.00   & 7949.97 \\
        LKH-3                      & 23.16                          & 0.17                         & 22.01  & 32.64  & 0.49   & 79.75   \\
        OR-Tools                   & 24.23                          & 4.82                         & 104.34 & 34.04  & 4.82   & 532.14   \\
        POMO                       & 30.52                          & 32.01                        & 4.28   & 46.49  & 43.15  & 35.89    \\
        DRL-2opt                   & 37.90                          & 63.93                        & 55.56  & 115.59 & 255.92 & 827.43  \\
        \begin{tabular}{@{}c@{}}Att-GCN \\ +MCTS\end{tabular}               & 23.86                          & 3.22                         & 5.85   & 33.42  & 2.91   & 200.28   \\
        \name                      & 24.65                          & 6.62                         & \textbf{0.33}    & 34.88  & 7.39\   & \textbf{0.72}     \\ \bottomrule
    \end{tabular}}
    \vspace{5pt}

    \setlength{\tabcolsep}{1mm}{
    \begin{tabular}{@{}lllllll@{}}
        \toprule
        \multirow{3}{*}{Algorithm} & \multicolumn{3}{l}{Random5000} & \multicolumn{3}{l}{Random10000}                                                \\ \cmidrule(l){2-7}
                                   & Length                         & Gap                             & Time      & Length  & Gap       & Time       \\ &                           & (\%)                            & (s)      &   & (\%)       & (s)       \\ \midrule
        LKH-3                      & 51.36                          & 0.00                          & 561.74  & 72.45   & 0.00    & 4746.59  \\
        OR-Tools                   & 53.35                          & 3.86                          & 5368.24 & 74.95   & 3.44    & 21358.66 \\
        POMO                       & 80.79                          & 57.29                         & 575.63  & OOM     & OOM       & OOM        \\
        DRL-2opt                   & 754.91                         & 1369.76                       & 2308.48 & 2860.86 & 3848.66 & 6073.43 \\
        \begin{tabular}{@{}c@{}}Att-GCN \\ +MCTS\end{tabular}               & 52.83                          & 2.86                          & 377.47  & 74.93   & 3.42    & 395.85   \\
        H-TSP                      & 55.01                          & 7.10                          & \textbf{1.66}    & 77.75   & 7.32    & \textbf{3.32}     \\ \bottomrule
    \end{tabular}}
    \vspace{-17pt}
\end{table}

\begin{figure}
    \centering
    \includegraphics[width=0.7\linewidth]{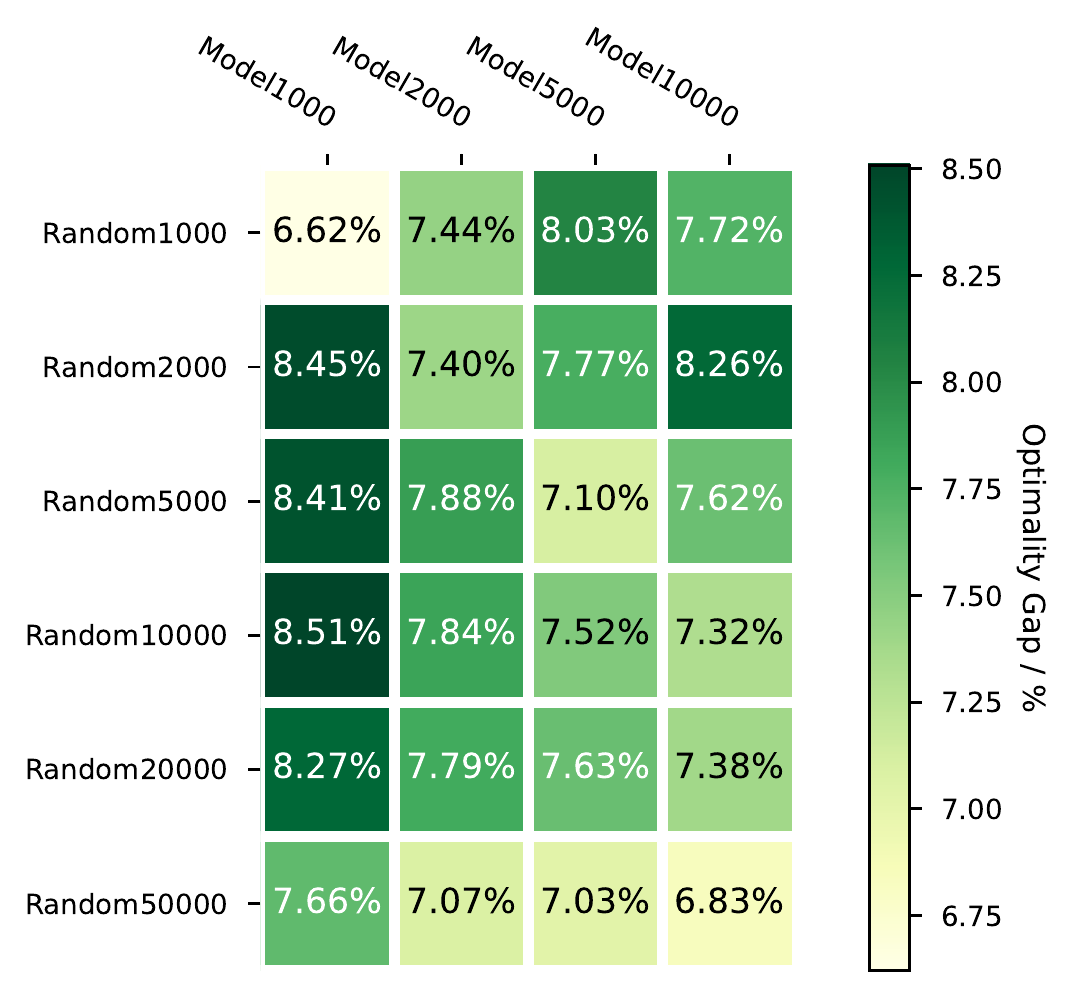}
    \caption{The optimality gaps of models tested on datasets out of training distribution}
    \label{fig:exp-generalization}
    \vspace{-17pt}
\end{figure}

Furthermore, we trained four models with different scales of TSP instances, denoted as Model1000, Model2000, Model5000 and Model10000, respectively.
In order to demonstrate the generalization ability of H-TSP,
we test the four trained models on randomly generated datasets with different number of nodes.
Figure~\ref{fig:exp-generalization} shows that H-TSP has a good generalization performance with TSP instances
from 1000 nodes to 50000 nodes.
Note that the optimality gap on Random50000 is smaller than the gap of Random20000,
because the optimal solutions of these datasets are generated by LKH-3,
and the solution quality of LKH-3 also declines as the number of nodes increases.

\vspace{-7pt}
\subsection{Ablation Study}
\begin{figure}
    \vspace{8pt}
    \includegraphics[width=\linewidth]{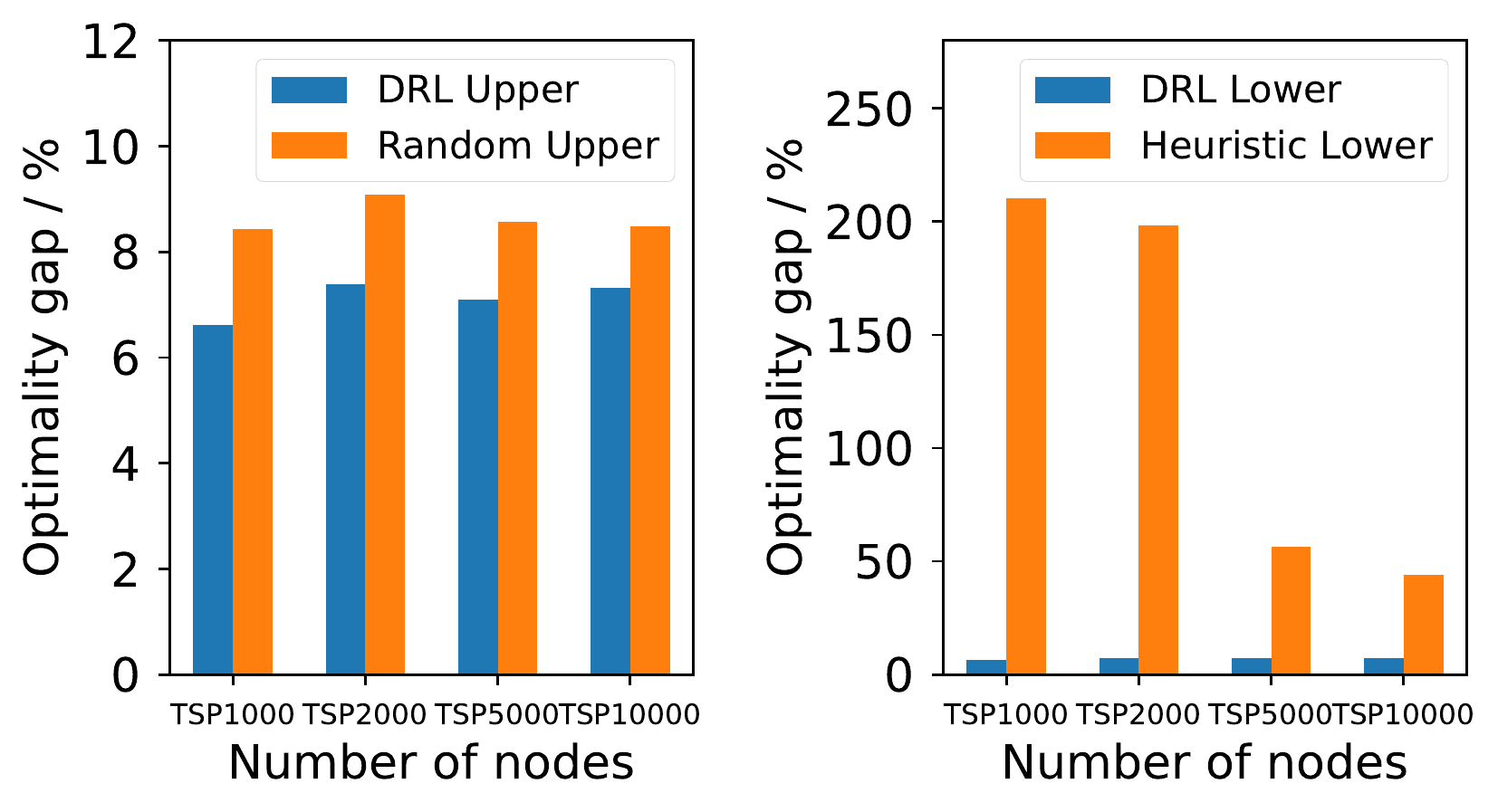}
    \caption{Optimality gap of models with different upper-level and lower-level model}
    \label{fig-ablation-upper-level-lower-level}
\end{figure}

\paragraph{Effect of two levels of models}
The H-TSP framework consists of two levels of models,
and both models are trained by DRL algorithms.
We conduct an ablation study during the inference stage to demonstrate the effect of each level
by replacing each of them with other heuristic methods.
For the upper-level model, we simply use a random policy to generate coordinates, starting which a sub-problem can be generated.
For the lower-level model, we introduce a simple but effective constructive heuristic called Farthest Insertion~\cite{rosenkrantz1974approximate}. 
We evaluate four combinations on the four datasets. The experimental results in Figure~\ref{fig-ablation-upper-level-lower-level} show that
our DRL-based framework outperforms other alternatives in all experiments.
Moreover, the effect of lower-level model is more significant than the upper-level model. 


Thus we further investigate this effect by combining LKH-3 with H-TSP, which is one of the SOTA heuristic TSP solvers.
LKH-3 is capable of solving numerous variants of TSP including our sub-problem.
We simply combine the trained upper-level model and the LKH-3 solver as the lower-level model.
The experimental results are shown in Table~\ref{tab:h-tsp-lkh}.
H-TSP with LKH-3 achieves better solution quality than the original H-TSP,
and it also takes longer solving time as LKH-3 is a search-based method that consumes more time than constructive methods.
However, if compared with Att-GCN+MCTS, H-TSP with LKH-3 still has higher efficiency, especially on larger TSPs.
This indicates that we can use the DRL-based lower-level model to speed up the training speed
and replace it with LKH-3 solver during inference, which makes H-TSP of better practical value.

\begin{table}[!h] \small
    \centering
    \vspace{-5pt}
    \caption{Comparison of H-TSP and LKH-3 solver }
    \vspace{-5pt}
    \label{tab:h-tsp-lkh}
    \begin{tabular}{@{}lllll@{}}
    \toprule
    \multirow{3}{*}{Algorithm}     & \multicolumn{2}{c}{Random1000} & \multicolumn{2}{c}{Random2000} \\ \cmidrule(l){2-5} 
                     & Gap           & Time           & Gap            & Time           \\ 
                     & (\%)          &(s)             & (\%)           &(s)             \\ \midrule
    H-TSP            & 6.62        & 0.33          & 7.39         & 0.72          \\
    H-TSP with LKH-3 & 4.06        & 3.53          & 5.01         & 6.88          \\
    Att-GCN+MCTS     & 3.22        & 5.85          & 2.91         & 200.28        \\ \hline
    \\

    \hline
    \multirow{3}{*}{Algorithm}     & \multicolumn{2}{c}{Random5000} & \multicolumn{2}{c}{Random10000} \\ \cmidrule(l){2-5} 
                     & Gap           & Time           & Gap            & Time           \\ 
                     & (\%)          &(s)             & (\%)           &(s)             \\ \midrule
    H-TSP            & 7.10        & 1.66          & 7.32         & 3.32          \\
    H-TSP with LKH-3 & 5.10        & 15.12         & 5.57         & 27.94         \\
    Att-GCN+MCTS     & 2.86        & 377.47        & 3.42         & 395.85        \\ \bottomrule
    \end{tabular}
    \vspace{-17pt}
\end{table}

\paragraph{Sub-problem generation and training strategy}
We conduct four ablation experiments on the sub-problem generation process and our training strategy.
"Visited fragment" refers to the fragment of visited nodes in sub-problem generation,
and "$k$-NN" means the $k$-NN graph.
“Joint training” and "warm-up" are the previously proposed training strategies.
Due to time limitations, we train the five models with TSP instances of 1000 nodes for 250 epochs,
and test them on Random1000. Table~\ref{tab:ablation-sub-prob-training} shows the test results,
where all four techniques are helpful in improving overall performance.
The warm-up stage of lower-level model has the greatest influence on the performance,
as the training of models on both levels become slow and unstable if starting with a poor lower-level policy.

\begin{table}[!h]
  \vspace{-5pt}
    \centering
    \caption{Analysis of four critical techniques}
    \label{tab:ablation-sub-prob-training}
    \begin{tabular}{@{}lll@{}}
    \toprule
                         & Gap (\%)            & $\Delta$ Gap (\%) \\ \midrule
    H-TSP                & \textbf{6.76} & 0.00                    \\
    w/o visited fragment & 7.42          & +0.66                   \\
    w/o $k$-NN           & 7.69          & +0.93                    \\
    w/o joint training   & 7.53          & +0.77                    \\
    w/o warm-up          & 27.05        & +20.29                   \\ \bottomrule
    \end{tabular}
    \vspace{-14pt}
\end{table}